\documentclass{article}

\PassOptionsToPackage{numbers, compress}{natbib}
\usepackage[final]{neurips_2023}

\usepackage[utf8]{inputenc} % allow utf-8 input
\usepackage[T1]{fontenc}    % use 8-bit T1 fonts
\usepackage{hyperref}       % hyperlinks
\usepackage{url}            % simple URL typesetting
\usepackage{booktabs}       % professional-quality tables
\usepackage{amsfonts}       % blackboard math symbols
\usepackage{nicefrac}       % compact symbols for 1/2, etc.
\usepackage{microtype}      % microtypography
\usepackage{xcolor}         % colors
\usepackage{amsmath}
\usepackage{graphicx}
\usepackage{todonotes}
\usepackage{caption}
\usepackage{subcaption}
\usepackage[title]{appendix}
\usepackage{bm}
\usepackage{algorithm2e}
\usepackage{listings,newtxtt}

\definecolor{deepblue}{rgb}{0,0,0.6}
\definecolor{deepred}{rgb}{0.6,0,0}
\definecolor{deepgreen}{rgb}{0,0.5,0}

\lstdefinestyle{lststyle}{
language=Python,
basicstyle=\ttfamily\small,
commentstyle=\color{deepred},
otherkeywords={self},             % Add keywords here
keywordstyle=\color{deepgreen},
emphstyle=\color{deepblue},    % Custom highlighting style
frame=tb,                         % Any extra options here
showstringspaces=false            % 
}

\newcommand{\ra}[1]{\renewcommand{\arraystretch}{#1}}

\title{Diffusion models for probabilistic programming}

\author{%
  Simon Dirmeier\thanks{Correspondence to: \texttt{simon.dirmeier@sdsc.ethz.ch}}\\
  Swiss Data Science Center\\
  ETH Zurich, Switzerland
  \And
  Fernando Perez-Cruz\\
  Swiss Data Science Center\\
  ETH Zurich, Switzerland
}

\begin{document}

\maketitle

\begin{abstract}
We propose \textit{diffusion model variational inference} (DMVI), a novel method for automated approximate inference in probabilistic programming languages (PPLs). DMVI utilizes diffusion models as variational approximations to the true posterior distribution by deriving a novel bound to the marginal likelihood objective used in Bayesian modelling. DMVI is easy to implement, allows hassle-free inference in PPLs without the drawbacks of, e.g., variational inference using normalizing flows, and does not make any constraints on the underlying neural network model. We evaluate DMVI on a set of common Bayesian models and show that its posterior inferences are in general more accurate than those of contemporary methods used in PPLs while having a similar computational cost and requiring less manual tuning.
\end{abstract}

\section{Introduction}
Probabilistic programming languages (PPLs) are computational tools that use inferential algorithms to automatically, i.e., without much user input, infer the posterior distributions of probabilistic models \cite{van2018introduction}. Due to their automated nature, PPLs have become an instrumental tool in applied sciences, such as computational physics and computational biology, by that democratizing the application of probabilistic machine learning and Bayesian statistics outside the circles of experts. Modern PPLs primarily use Markov chain Monte Carlo (MCMC, \cite{brooks2011handbook}) or optimization-based methods for probabilistic inference. While MCMC methods, such as Hamiltonian Monte Carlo (HMC, \cite{betancourt2017conceptual}), can yield highly accurate posterior inferences and work well off-the-shelf for a broad class of models, they can be laboriously slow, for example with high sample sizes, large parameter dimensionalities, multi-modal posteriors, or when the posterior geometry suffers from extreme curvature. Optimization-based methods such as variational inference using normalising flows (NFVI, \cite{blei2017var,rezende15nfvi}) or automatic differentiation variational inference (ADVI, \cite{kucukelbir2017automatic}) instead approximate the posterior distribution by utilizing a trainable distribution and optimizing its parameters such that it resembles the target distribution closely. However, the accuracy of simple methods like ADVI can suffer in complex models due to its reliance on simple exponential familes as variational guides. NFVI, on the other hand, allows for highly complex approximations by stacking several normalizing flow (NF) layers that increase the expressivity of the posterior approximation. A drawback of NFVI is that it requires the users to have advanced understanding of both the problem to be modelled as well as NF architecture to be used and how it can be optimized in an error-free manner. For instance, off-the-shelf NF architectures which are commonly found in PPLs can be numerically unstable using 32-bit floating point arithmetic. NFVI furthermore puts architectural constraints on the neural networks to be used which reduces the number of available architectures. For instance, inverse autoregressive flows (IAFs, \cite{kingma2016improved}), one of the pre-dominant NFVI approaches, uses MADE neural networks \citep{germain15made} at their core to ensure that the variables to be modelled factor autoregressively such that an efficient computation of the density of a data point can be guaranteed. 

In this work, we introduce \textit{diffusion model variational inference} (DMVI) which uses recent advances in diffusion probabilistic modelling to derive a new objective for variational inference. In comparison to NFVI, DMVI does not have any architectural constraints such that any neural network can be used. We show that off-the-shelf, i.e., without architectural considerations for the score model and no user input, DMVI achieves state-of-the-art performance on several benchmark models. Since DMVI requires to iterate over a reverse diffusion process to generate samples which slows down sampling tremendously, we make use of an efficient sampling technique from the recent literature which reduces both training and sampling time to similar scales as NFVI.

\section{Background}

Diffusion probabilistic models (DPMs, \cite{sohldickstein15deep,ho2020diffusion,song2019generative}) are latent variable models of the form:
\begin{equation}
    p_{\phi}(\bm{y}_{0}) = \int p_{\phi}(\bm{y}_{0:T}) d\bm{y}_{1:T} =  \int  p(\bm{y}_{T})  \sum_{t=1}^T p_{\phi}(\bm{y}_{t - 1} | \bm{y}_{t}) d\bm{y}_{1:T}
\label{eqn:reverse-process}
\end{equation}
Equation~\ref{eqn:reverse-process} is called \textit{reverse process}, the transitions $p_{\phi}(\bm{y}_{t - 1} | \bm{y}_{t})$ are parameterized by a \textit{score model} with neural network weights $\phi$ (see \cite{ho2020diffusion,song2019generative,kingma2021variational} for denotation), and $T$ is the number of diffusion steps. DPMs define a complementary \textit{forward process} starting from $\bm{y}_0 \sim r(\bm{y}_0)$ as $r(\bm{y}_{1:T} | \bm{y}_0) = \prod_{t=1}^T r(\bm{y}_t | \bm{y}_{t - 1})$, such that the conditional distribution of any intermediate random variable can be represented as $r(\bm{y}_t | \bm{y}_{0} ) = \mathcal{N}(\bm{y}_t; \alpha_t \bm{y}_{0}, \sigma^2_t \bm{I})$ where $\sigma^2_t$ is a pre-defined variance schedule and $\alpha_t = \sqrt{1 - \sigma^2_t}$. Training of the neural network parameters is performed by maximizing the evidence lower bound
\begin{equation*}
\mathbb{E}_{r} \biggl[ 
    \log p_\phi \left(\bm{y}_0 | \bm{y}_1 \right) -
    \sum_{t=2}^T \mathbb{KL}\Bigl[ r(\bm{y}_{t - 1} | \bm{y}_{t}, \bm{y}_0), p_\phi(\bm{y}_{t - 1} | \bm{y}_t)  \Bigr]  -
    \mathbb{KL}\Bigl[ r(\bm{y}_T | \bm{y}_0), p_\phi(\bm{y}_T)  \Bigr]
    \biggr]
\end{equation*} 
where the forward process posterior $r(\bm{y}_{t - 1} | \bm{y}_{t}, \bm{y}_0)$ can be computed analytically. Ho et al. \cite{ho2020diffusion} derive a simplified objective that improves sample quality and can enhance numerical stability (see Appendix~\ref{appendix:more-background}). This objective above avoids evaluating the entire forward process during, since only a single sample $\bm{y}_t$ from the variational posterior needs to be drawn per train step. Sampling $\bm{y}_0$ from a trained model, however, requires traversing the entire chain $p_{\phi}(\bm{y}_{0:T})$. To speed up this process, Lu et al. \cite{lu2022dpmsolver} propose an efficient ODE-solver that can generate high-quality samples in only 10-20 steps which we use during training and sampling.

\section{Diffusion model variational inference}

We introduce a novel approach for automated variational inference (VI) for probabilistic programming languages which we term \textit{diffusion model variational inference} (DMVI). 

We model the variational approximation $q(\boldsymbol \theta)$ to the posterior $p(\boldsymbol \theta | \bm{y})$ of a Bayesian model using a DPM by applying the variational principle to the marginal likelihood twice (c.f. \cite{ranganath16hierarchical}) and derive the objective
\begin{equation}
\begin{aligned}
\log p(\bm{y}) & \ge \mathbb{E}_{q(\boldsymbol \theta)} \left[\log p(\bm{y}, \boldsymbol \theta) - \log q(\boldsymbol \theta) \right]  \\ 
& = \mathbb{E}_{q(\boldsymbol \theta)} \left[\log p (\bm{y}, \boldsymbol \theta) - \log  \mathbb{E}_{r(\bm{w}_{1:T} |  \boldsymbol\theta)} \left[  \dfrac{q(\boldsymbol \theta, \bm{w}_{1:T} )}{r(\bm{w}_{1:T} | \boldsymbol\theta)}  \right] \right]  \\
& \ge  \mathbb{E}_{q(\boldsymbol \theta)} \left[\log p(\bm{y}, \boldsymbol \theta) - \mathbb{E}_{r(\bm{w}_{1:T}|  \boldsymbol\theta)} \left[ \log  \dfrac{q(\boldsymbol \theta, \bm{w}_{1:T} )}{r(\bm{w}_{1:T}| \boldsymbol\theta)}  \right] \right]\\ 
& = \mathbb{E}_{q(\boldsymbol \theta), r(\bm{w}_{1:T}| \boldsymbol\theta)}\left[\log p (\bm{y}, \boldsymbol \theta) - \log \dfrac{q(\boldsymbol \theta, \bm{w}_{1:T} )}{r(\bm{w}_{1:T}|  \boldsymbol\theta)}\right]
\end{aligned}
\label{eqn:dmvi}
\end{equation}
where we for notational convenience drop the parameters of the guide (see Appendix~\ref{appendix:additional-math} for a detailed derivation and Appendix~\ref{appendix:more-background} for additional background on VI). Equation~\eqref{eqn:dmvi} models the distribution $q(\boldsymbol \theta)$ using a diffusion model that is defined via the reverse process $q(\boldsymbol \theta, \bm{w}_{1:T}) = q_{\phi}(\boldsymbol \theta | \bm{w}_{1}) \prod_{t=2}^T q_{\phi}(\bm{w}_{t - 1} | \bm{w}_{t}) q(\bm{w}_{T})$ following the derivation in Equation~\eqref{eqn:reverse-process} and the complementary forward process $r(\bm{w}_{1:T} | \boldsymbol \theta)$ (see Algorithm~\ref{alg:dmvi} and Appendix~\ref{appendix:implementation} for implementation details). For constrained parameters $\boldsymbol \theta$, we follow the same approach as ADVI and transform the parameters into an unconstrained space via a bijection $f$ as $\boldsymbol {\xi} \leftarrow f(\boldsymbol \theta)$, model the distribution of $\boldsymbol {\xi}$ using a diffusion model in that space, and apply the inverse transformation to parameterize $p(\bm{y}, \boldsymbol \theta)$ (see Appendix~\ref{appendix:additional-math}).

\RestyleAlgo{ruled}
\begin{algorithm}[h!t]
\caption{DMVI}
    
\SetKwInput{Outputs}{Outputs}
\SetKwInput{Inputs}{Inputs}
\SetKwInput{Initialize}{Initialize}
\Inputs{data set $\mathcal{D}$, joint distribution $p(\bm{y}, \boldsymbol \theta)$ (optionally with generative parameters $\psi$), score model with variational parameters $\phi$}
\While{not converged}{
$\bm{y}$ $\leftarrow$ sample mini-batch from data set $\mathcal{D}$\\
$\boldsymbol \theta, \bm{w} \sim q(\boldsymbol \theta, \bm{w}_{1:T})$ using DPM-solver\\
$\text{evidence}(\bm{y}) \leftarrow$ evaluate Eqn.~\eqref{eqn:dmvi} \\
$\Delta \phi \propto - \nabla_{\phi} \text{evidence}(\bm{y})$\\
(optionally: $\Delta \psi  \propto - \nabla_{\psi} \text{evidence}(\bm{y})$)\\
}
\label{alg:dmvi}
\end{algorithm}

\section{Experiments}

We evaluate DMVI on three generative models, a Gaussian mean model, a hierarchical model and a multivariate Gaussian mixture model with different sample sizes and a variety of hyper-parameter settings, and compare it to NFVI and mean-field ADVI. Briefly, we train DMVI with different numbers of total diffusions steps ($N_\text{d} = 50, 100$), and different numbers of DPM-Solver order and steps ($N_\text{o} = 1, 3$ and $N_\text{s} = 10, 20$; see \cite{lu2022dpmsolver} for details). We evaluate the performance of each method by computing the mean squared error (MSE) between a posterior sample of size $20\, 000$ of a method and the prior parameter configuration that was used to simulate synthetic data set of size $N = 100, 1000$ from a generative model. Furthermore, we evaluate both training time ($\bar{T}_\text{train}$) and sampling time ($\bar{T}_\text{sample}$) since these are often decisive factors which inferential algorithm is chosen by a user of a PPL, e.g., for quick prototyping and model checking. We replicate each experiment $5$ times with different random number generation seeds and report the averages of the three aforementioned metrics over these runs. Full experimental details and source code for reproducibility can be found in Appendix~\ref{appendix:experiments} or GitHub, respectively. More experimental results can be found in Appendix~\ref{appendix:more-experiments}.

\paragraph{Mean model}

We first evaluate DMVI on the following simple generative model:
\begin{equation}
\bm{y}_n \sim \text{MvNormal}(\boldsymbol \mu, \bm{I}), \quad \boldsymbol \mu \sim \text{MvNormal}(\bm{0}, \bm{I})  \qquad \forall n = 1, \dots, N
\end{equation}
For this model, neither NFVI nor DMVI should significantly outperform a simple approach like ADVI. Indeed the methods perform comparably w.r.t. the MSE, but curiously DMVI has a minor performance advantage over NFVI and ADVI (Table~\ref{table:evaluated-models}). However in all cases the average training times $\bar{T}_\text{train}$ and sampling times $\bar{T}_\text{sample}$ are not competitive despite using the DPM-solver implementation \citep{lu2022dpmsolver}.

\paragraph{Hierarchical model}

We next evaluate DMVI on a more interesting model, i.e., a two-level hierarchical model of the following form:
\begin{equation}
\begin{split}
\gamma_i & \sim  \text{Normal}(\mu_\gamma, \sigma^2_\gamma), \quad \mu_\gamma \sim \text{Normal}(0, 1), \sigma_\gamma \sim \text{HalfNormal}(1) \qquad \forall i = 1, \dots, 5 \\
\beta_{ij} & \sim  \text{Normal}(\gamma_i, \sigma^2_\beta), \quad \sigma_\beta  \sim  \text{HalfNormal}(1) \qquad \forall j = 1, 2  \\
y_{ijn}     & \sim \text{Normal}(\beta_{ij}, 1) \qquad \forall n = 1, \dots, N
\end{split}
\end{equation}
The statistical dependencies of the model are difficult to resolve and need to be learned during training. While mean-field ADVI can not account for the correlation structure, NFVI and DMVI can learn them directly from data. The constrained parameters of the model induce a pathological posterior geometry that can pose a significant challenge even to HMC methods and which becomes more prominent with increased sample sizes. For this model DMVI has a significant performance advance over ADVI and NVFI (Table~\ref{table:evaluated-models}). Furthermore, some DPM-Solver parameterizations also show competitive average training and sampling times in comparison to NFVI for both sample sizes, e.g., using $N_\text{steps} = 10$ and $N_\text{order} = 3$.

\paragraph{Mixture model}

Finally, we evaluate DMVI on a bivariate Gaussian mixture model with $K=3$ components:
\begin{equation}
\begin{split}
\mu_{ki} & \sim  \text{Normal}(0, 1), \quad \sigma_{ki}  \sim  \text{HalfNormal}(1) \qquad\forall k = 1, 2, 3 \quad \forall i = 1, 2  \\
y_{n}         & \sim \prod_{k=1}^K \pi_k \text{MvNormal}(\boldsymbol \mu_k, \boldsymbol \Sigma_k)\,\; \quad \forall n = 1, \dots, N
\end{split}
\end{equation}
where $\boldsymbol \pi = \text{Dirchlet}\left(1, 1\right)$, $\boldsymbol \mu_k = (\mu_{k1}, \mu_{k2})^T$ and $\boldsymbol \Sigma_k = \text{diag}(\boldsymbol \sigma^2_{k})$ are diagonal covariance matrices. Despite setting the mixing weights, the model is non-identifiable. For a sample size of $N=100$ DMVI outperforms ADVI significantly while being on par with NFVI (Table~\ref{table:evaluated-models}). As expected, increasing the sample size to $N=1000$ reduces the error of ADVI where it has a minor advantage over both DMVI and NFVI. Both average training and sample are competitive w.r.t. NFVI or even outperforming it.

\begin{table*}[h!]
\centering
\ra{1}
\begin{tabular}{@{}r rr rrr rrr c rrr@{}}\toprule
& & &&& \multicolumn{3}{c}{$N=100$} & \phantom{} & \multicolumn{3}{c}{$N=1000$} \\
\cmidrule{6-8} \cmidrule{10-12} 
       &       & $N_\text{d}$ & $N_\text{s}$ & $N_\text{o}$ & $\bar{T}_\text{train}$ & $\bar{T}_\text{sample}$ & MSE && $\bar{T}_\text{train}$ & $\bar{T}_\text{sample}$ & MSE\\
\midrule

Mean   &ADVI            &&&& $5.64$ & $0.28$ & $0.11$ && $19.59$ & $0.37$ & $0.06$\\
model  & DMVI &$50$ & $10$ & $1$ & $42.16$ & $4.00$ & ${0.07}$ && $270.94$ & $4.05$ & $0.06$\\
        &&$50$ & $10$ & $3$ & $53.66$ & $7.35$ & $0.07$ && $256.45$ & $7.22$ & ${0.06}$\\
        &&$50$ & $20$ & $1$ & $72.05$ & $6.24$ & $0.07$ && $565.43$ & $6.90$ & $0.06$\\
        &&$50$ & $20$ & $3$ & $75.19$ & $10.39$ & $0.07$ && $429.22$ & $10.63$ & $0.06$\\

        &&$100$ & $10$ & $1$ & $54.52$ & $5.88$ & $0.07$ && $353.74$ & $5.64$ & $0.06$\\
        &&$100$ & $10$ & $3$ & $60.93$ & $9.43$ & $0.07$ && $323.14$ & $8.88$ & $0.06$\\
        &&$100$ & $20$ & $1$ & $92.94$ & $9.36$ & $0.07$ && $700.45$ & $10.21$ & $0.06$\\
        &&$100$ & $20$ & $3$ & $96.60$ & $14.23$ & $0.07$ && $607.68$ & $15.82$ & $0.06$\\

        &NFVI             &&&& ${35.34}$ & ${1.13}$ & $0.09$ && ${196.55}$ & ${1.05}$ & $0.07$\\

\midrule
Hierarchical & ADVI &&&& $7.38$ &	$0.37$ & $11.77$                     && $38.87$ & $0.40$ & $2.54$\\
model        & DMVI &$50$ & $10$ & $1$ & $57.18$ &	$4.69$  & $0.49$                 && $276.22$ &	$3.93$ & $1.09$\\
             &      &$50$ & $10$ & $3$ & ${54.52}$  & $7.12$ & $0.46$        && ${266.37}$ &	$7.23$ & $1.07$\\
             &&$50$ & $20$ & $1$ & $92.67$ &	$7.70$  & $0.51$                 && $546.03$ &	$6.85$ & $1.10$\\
             &&$50$ & $20$ & $3$ & $83.29$ &	$10.83$ & $0.52$                 && $491.99$ &	$11.22$ & $1.05$\\

            &&$100$ & $10$ & $1$ & $66.07$  & $6.29$   & $0.41$             && $437.52$ & $6.27$  & $2.22$\\
            &&$100$ & $10$ & $3$ & $64.50$  & $9.22$   & $\mathbf{0.39}$    && $353.74$ & $9.18$ & $\mathbf{1.03}$\\
            &&$100$ & $20$ & $1$ & $110.28$ & $10.55$  & $0.40$             && $757.28$ & $9.78$  & $3.47$\\
            &&$100$ & $20$ & $3$ & $107.82$ & $15.53$  & $0.44$             && $605.38$ & $14.38$ & $1.03$\\

            &NFVI &&&& $96.01$ 	& ${1.17}$ &$0.62$                     && $689.78$ & ${1.32}$ & $3.99$\\

\midrule
Mixture & ADVI &      &      &     &  $6.84$ & $0.35$ & $8.49$                   && $23.72$ & $0.37$ & $\mathbf{0.75}$\\
model   & DMVI & $50$ & $10$ & $1$ & ${51.37}$ & $4.33$ & $\mathbf{0.86}$ && $297.47$ & $4.01$ & $0.77$\\
        &      & $50$ & $10$ & $3$     & $57.43$ & $7.70$ & $0.89$               && ${288.42}$ & $7.48$ & ${0.77}$\\
        &      & $50$ & $20$ & $1$     & $77.23$ & $6.34$ & $0.87$               && $556.48$ & $6.35$ & $0.77$ \\
        &      & $50$ & $20$ & $3$     & $91.47$ & $11.75$ & $0.90$              && $473.91$ & $11.16$ & $0.77$\\
        
        &      & $100$ & $10$ & $1$    & $55.81$ & $5.78$ & $0.88$               && $361.54$ & $5.63$ & $0.77$\\
        &      & $100$ & $10$ & $3$    & $69.59$ & $9.52$ & $0.88$               && $306.45$ & $8.47$ & $0.77$\\
        &      & $100$ & $20$ & $1$    & $99.82$ & $9.86$ & $0.89$               && $639.02$ & $9.03$ & $0.77$\\
        &      & $100$ & $20$ & $3$    & $96.03$ & $13.98$ & $0.89$              && $564.77$ & $13.37$ & $0.77$\\

       & NFVI &        &      &        & $77.34$ & ${1.26}$ & $0.89$      && $440.61$ & ${1.17}$ & $0.85$\\
\bottomrule
\end{tabular}
\caption{Evaluated models (lower is better).}
\label{table:evaluated-models}
\end{table*}

\section{Conclusion}

We presented {diffusion model variational inference}, a novel approach for automated approximate inference in PPLs. DMVI achieves state-of-the-art performance on several experimental models and is generally on par with NFVI w.r.t. required computational resources.

We designed DMVI with the goal in mind to introduce a novel method that requires less expertise in probabilistic deep learning and thus open PPLs up for broader user bases. By that we possibly trade simplicity for inferential accuracy and increased training and sampling times. 
DMVI does not have any architectural constraints on the neural network model which reduces the complexity of designing guides for the user and which, for instance, allows to easily incorporate normalization layers such as BatchNorm or LayerNorm to reduce numerical instabilities. 

Our work is a first step to establish diffusion models for variational approximate inference and we hope that it will open up interesting avenues for future research.

\newpage
\bibliographystyle{plain}
\bibliography{references} 

\newpage
\begin{appendices}
\section{Background}
\label{appendix:more-background}
\subsection{Variational inference}

We are interested in inference of the posterior distribution $p(\boldsymbol \theta | \bm{y}) \propto p(\bm{y} | \boldsymbol \theta) p(\boldsymbol \theta)$ for a parameter $\boldsymbol \theta$ and data set $\bm{y}$. In cases where the posterior is not analytically available, in variational inference we approximate it using some variational distribution, also called guide, $q_\psi(\boldsymbol \theta)$ and optimize a lower bound to the marginal likelihood $p(\bm{y})$

\begin{equation}
\log p(\bm{y}) \ge \mathbb{E}_{q_\psi(\boldsymbol \theta)} \left[\log p(\bm{y}, \boldsymbol \theta) - \log q(\boldsymbol \theta) \right] 
\end{equation}

with respect to the variational parameters $\psi$. In probabilistic programming, methods of choice include ADVI \citep{kucukelbir2017automatic} where the guides are chosen to be unconstrained Gaussians that are transformed to the domain space of the posterior parameters, or NFVI \citep{rezende15nfvi}, in particular using inverse autoregressive flows (IAFs, \cite{kingma2016improved}), where the guides are parameterized by trainable bijections and which allow to sample from the variational guide and evaluate the log-probability of a sample efficiently.

\subsection{Diffusion probabilistic models}
\label{appendix:dpms}
Ho et al. \cite{ho2020diffusion} derive a simplified objective that improves sample quality and can enhance numerical stability 

\begin{equation*}
L := \mathbb{E}_{r} \Bigl[ \lVert \boldsymbol  \epsilon - \boldsymbol \epsilon_{\phi} (\alpha_t \bm{y}_0 + \sigma_t \boldsymbol \epsilon, t) \rVert^2 \Bigr]
\end{equation*} 

where $\boldsymbol \epsilon \sim \mathcal{N}(\bm{0}, \bm{I})$ and $\boldsymbol \epsilon_\phi(\cdot, t)$ is a neural network that aims to predict the noise that perturbed the sample $\bm{y}_0$ at time step $t$. The objective above avoids evaluating the entire forward process during training, since only a single sample $\bm{y}_t$ from the variational posterior needs to be drawn per train step. 

\section{Mathematical derivations}
\label{appendix:additional-math}

\subsection{Derivation of lower bound to evidence lower bound}

Our novel objective uses a lower bound to the ELBO. Its derivation is as follows:

\begin{equation}
\begin{split}
\log  p(\bm{y}) & \ge \mathbb{E}_{q(\boldsymbol \theta)} \left[\log p(\bm{y}, \boldsymbol \theta) - \log q(\boldsymbol \theta) \right]  \\
& = \mathbb{E}_{q(\boldsymbol \theta)} \left[\log p(\bm{y}, \boldsymbol \theta) - \log  \int q(\boldsymbol \theta, \bm{w}_{1:T}) d\bm{w} \right]  \\
& = \mathbb{E}_{q(\boldsymbol \theta)} \left[\log p(\bm{y}, \boldsymbol \theta) - \log \left[ \int q(\boldsymbol \theta, \bm{w}_{1:T}) \frac{r(\bm{w}_{1:T}|  \boldsymbol\theta)}{r(\bm{w}_{1:T}|  \boldsymbol\theta)} d\bm{w}\right] \right]  \\
& = \mathbb{E}_{q(\boldsymbol \theta)} \left[\log p (\bm{y}, \boldsymbol \theta) - \log  \mathbb{E}_{r(\bm{w}_{1:T} |  \boldsymbol\theta)} \left[  \dfrac{q(\boldsymbol \theta, \bm{w}_{1:T} )}{r(\bm{w}_{1:T} | \boldsymbol\theta)}  \right] \right]  \\
& \ge  \mathbb{E}_{q(\boldsymbol \theta)} \left[\log p(\bm{y}, \boldsymbol \theta) - \mathbb{E}_{r(\bm{w}_{1:T}|  \boldsymbol\theta)} \left[ \log  \dfrac{q(\boldsymbol \theta, \bm{w}_{1:T} )}{r(\bm{w}_{1:T}| \boldsymbol\theta)}  \right] \right]\\ 
& = \int q(\boldsymbol \theta) \left[ \log p (\bm{y}, \boldsymbol \theta) 
- \left[ \int r(\bm{w}_{1:T}) \left[ \log  \dfrac{q(\boldsymbol \theta, \bm{w}_{1:T} )}{r(\bm{w}_{1:T}| \boldsymbol\theta)}  \right] d \bm{w} \right] 
\right] d \boldsymbol \theta \\
& = \int  q(\boldsymbol \theta) \left[\int r(\bm{w}_{1:T}| \boldsymbol\theta) \log p (\bm{y}, \boldsymbol \theta) \ d \bm{w} \right] -\;\left[ \int r(\bm{w}_{1:T}| \boldsymbol\theta) \left[ \log  \dfrac{q(\boldsymbol \theta, \bm{w}_{1:T} )}{r(\bm{w}_{1:T} | \boldsymbol\theta)}  \right] d \bm{w} \right] 
d \boldsymbol \theta \\
& = \int \int q(\boldsymbol \theta)  r(\bm{w}_{1:T}|  \boldsymbol\theta) \left[\log p (\bm{y}, \boldsymbol \theta)  - \log \dfrac{q(\boldsymbol \theta, \bm{w}_{1:T} )}{r(\bm{w}_{1:T}| \boldsymbol\theta)}\right] d\bm{w} d\boldsymbol \theta  \\
& = \mathbb{E}_{q(\boldsymbol \theta), r(\bm{w}_{1:T}| \boldsymbol\theta)}\left[\log p (\bm{y}, \boldsymbol \theta) - \log \dfrac{q(\boldsymbol \theta, \bm{w}_{1:T} )}{r(\bm{w}_{1:T}|  \boldsymbol\theta)}\right]
\end{split}
\label{eqn:elboo}
\end{equation}

\subsection{Transformation of constrained variables}

Consider a constrained variable $\boldsymbol \theta \sim p(\boldsymbol \theta)$, i.e., a variable that does not have support on the real line. We transform $\boldsymbol \theta$ onto the real line using a bijection $f$

\begin{equation*}
\boldsymbol \xi \leftarrow f(\boldsymbol \theta)
\end{equation*}

which we in practice do using the \texttt{experimental\_default\_event\_space\_bijector} method of TensorFlow Probability \citep{dillon2017tensorflow} which automatically chooses an appropriate unconstraining bijection for a constrained parameter $\boldsymbol \theta$.

The joint density of parameter $\boldsymbol \xi$ and data $\bm{y}$ is then

\begin{equation*}
p(\bm{y}, \boldsymbol \xi) = p\left(\bm{y}, f^{-1}(\boldsymbol \xi)\right) \big| \det J_{f^{-1}}(\boldsymbol \xi) \big|
\end{equation*}

where $\det J_{f^{-1}}(\boldsymbol \xi)$ is the Jacobian determinant of the inverse transformation which is required to adjust for changes in volumne between the two joint densities. The above density is used as likelihood function within the ELBO above (c.f. \cite{kucukelbir2017automatic}). We use these kind of reparameterizations in all models with priors with constrained supports. The entire objective used for training then becomes

\begin{equation*}
\mathbb{E}_{q(\boldsymbol \xi), r(\bm{w}_{1:T}| \boldsymbol \xi)}\left[ \log p \left( \bm{y}, f^{-1} \left(\boldsymbol \xi \right) \right) + \log \big| \det J_{f^{-1}}(\boldsymbol \xi) \big|  - \log \dfrac{q(\boldsymbol \xi, \bm{w}_{1:T} )}{r(\bm{w}_{1:T} | \boldsymbol \xi)} \right]
\end{equation*}

which is amendable to optimization using stochastic variational inference \citep{hoffman2013stochastic}.

\section{Experimental details}
\label{appendix:experiments}

As reported before, we train DMVI with different numbers of total diffusions steps ($N_\text{d} = 50, 100$), and different numbers of DPM-Solver steps and order ($N_\text{o} = 1, 3$ and $N_\text{s} = 10, 20$; see \cite{lu2022dpmsolver} for details). 

We train each model until convergence on mini-batches of size $32$ using an AdamW optimizer \citep{loshchilov2018decoupled} for training  with a learning rate of $l=0.001$. To optimize the objective (Equation~\eqref{eqn:elboo}), we take Monte Carlo samples of size $5$ for all methods. ADVI and NFVI are trained in the same fashion. 

Each model uses the identical training routine (but a different variational guide) and is implemented using a custom JAX \cite{jax2018github} module to make the experimental training and posterior sampling times as comparable as possible. Each model has been evaluated on the same machine with identical computational resources (i.e., CPU and memory). We evaluate the performance of each method by computing the mean squared error (MSE) between a posterior sample of size $20\, 000$ of a method and the prior parameter configuration that was used to simulate synthetic data set of size $N = 100, 1000$ from a generative model. Each experiment is replicated $5$ times with different random number generation seeds and the averages of the three aforementioned metrics over these runs is reported. 

We use the same network architectures for each experiment and chose them somewhat arbitrarily without consideration for the complexity of the prior model of an experiment.

DMVI uses a simple MLP with one hidden layer of $256$ nodes as score model. We used \texttt{gelu} activation functions throughout. We use layer normalization and dropout of $0.1$ before projecting the hidden representation using a linear layer to the dimensionality of the parameter space. We use a linear noise schedule from $\beta_\text{min}=10^{-4}$ to $\beta_\text{max}=0.02$ (we found the cosine noise schedule of \cite{nichol21improved} to be numerically unstable in our experiments). We use the simplified objective derived by Ho et al. \cite{ho2020diffusion} (Appendix~\ref{appendix:dpms}) within Equation~\ref{eqn:elboo}.

NFVI uses an NF with three NF layers, consisting of an IAF layer with a 256 node MADE network, a permutation layer, and another IAF layer with a 256 node MADE network. NFVI also uses \texttt{gelu} activation functions. We initialized all weights to zero after we found that other initializations (such as truncated Normals with small standard deviation) yielded \texttt{NaN}s during the first steps of training.

Source code to reproduce all results can be found at \href{https://github.com/dirmeier/dmvi}{\url{https://github.com/dirmeier/dmvi}}.

\newpage
\section{Implementation details}
\label{appendix:implementation}

DMVI can be implemented in the same manner as ADVI and NFVI. Within a general-purpose probabilistic programming language (and for our experiments), one could design a modular framework by implementing a \texttt{Guide} abstract base class that exposes two public methods \texttt{sample} (which samples from the guide) and \texttt{evidence} (which evaluates the probability/evidence of a sample). Below we exemplify these implementations using the neural network library \texttt{Haiku} \cite{haiku2020github}. The base class looks as follows:\\

\begin{lstlisting}[emph={__init__,DDPM,Guide,sample,evidence,__call__,diffusion_loss}]
import abc
import jax

class Guide(metaclass=abc.ABCMeta):

    @abc.abstractmethod
    def evidence(self, theta) -> jax.Array:
        pass
        
    @abc.abstractmethod
    def sample(self, sample_shape=(1,)) -> jax.Array:
        pass
\end{lstlisting}

A DMVI guide could then be implemented as below:\\

\begin{lstlisting}[emph={__call__,DDPM,sample,evidence,__call__,diffusion_loss}]
import haiku as hk

class DDPM(Guide, hk.Module):    
    def __call__(self, method="evidence", **kwargs):
        return getattr(self, method)(**kwargs)

    def evidence(self, theta) -> jax.Array:
        # evaluate diffusion loss and return its negative
        obj = -self.diffusion_loss(theta)        
        return obj

    def diffusion_loss(self, theta) -> jax.Array:
        ...
        
    def sample(self, sample_shape=(1,)) -> jax.Array:
        # sample using DPM-Solver
        ...        
\end{lstlisting}

The guides of ADVI and NFVI are implemented analogously with the exception that \texttt{evidence} returns the "exact" log-probability of a parameter value.

\newpage
\section{More experimental results}
\label{appendix:more-experiments}

In this section, we provide several, additional experimental results. Particularly, we conducted more experiments on different parameterizations and graphical structures of the prior models of the hierarchical model from the main manuscript.

\subsection{Mean model}

\begin{equation}
\begin{split}
\boldsymbol \mu &\sim \text{MvNormal}(\bm{0}, \bm{I}) \\
\bm{y}_n &\sim \text{MvNormal}(\boldsymbol \mu, \bm{I}) \qquad \forall n = 1, \dots, N
\end{split}
\label{eqn:mean-model}
\end{equation}

\begin{table*}[h!]
\centering
\ra{1}
\begin{tabular}{@{}rrrrr rrr c rrr@{}}\toprule
&&&&& \multicolumn{3}{c}{$N=100$} & \phantom{abc} & \multicolumn{3}{c}{$N=1000$} \\
\cmidrule{6-8} \cmidrule{10-12} 
&$N_\text{diff}$ & $N_\text{steps}$ & $N_\text{order}$ && $\bar{T}_\text{train}$ & $\bar{T}_\text{sample}$ & MSE && $\bar{T}_\text{train}$ & $\bar{T}_\text{sample}$ & MSE\\
\midrule 
ADVI            &&&&& $5.64$ & $0.28$ & $0.11$ && $19.59$ & $0.37$ & $0.06$\\
DMVI&$50$ & $10$ & $1$ && $42.16$ & $4.00$ & ${0.07}$ && $270.94$ & $4.05$ & $0.06$\\
&$50$ & $10$ & $3$ && $53.66$ & $7.35$ & $0.07$ && $256.45$ & $7.22$ & ${0.06}$\\
&$50$ & $20$ & $1$ && $72.05$ & $6.24$ & $0.07$ && $565.43$ & $6.90$ & $0.06$\\
&$50$ & $20$ & $3$ && $75.19$ & $10.39$ & $0.07$ && $429.22$ & $10.63$ & $0.06$\\

&$100$ & $10$ & $1$ && $54.52$ & $5.88$ & $0.07$ && $353.74$ & $5.64$ & $0.06$\\
&$100$ & $10$ & $3$ && $60.93$ & $9.43$ & $0.07$ && $323.14$ & $8.88$ & $0.06$\\
&$100$ & $20$ & $1$ && $92.94$ & $9.36$ & $0.07$ && $700.45$ & $10.21$ & $0.06$\\
&$100$ & $20$ & $3$ && $96.60$ & $14.23$ & $0.07$ && $607.68$ & $15.82$ & $0.06$\\

NFVI             &&&&& ${35.34}$ & ${1.13}$ & $0.09$ && ${196.55}$ & ${1.05}$ & $0.07$\\
\bottomrule
\end{tabular}
\caption{Model of equation \ref{eqn:mean-model}.}
\end{table*}

\subsection{Mixture model}

\begin{equation}
\begin{split}
\mu_{ki}        & \sim  \text{Normal}(0, 1) \qquad \forall k = 1, \dots, 3 \qquad \forall i = 1, \dots, 2  \\
\sigma_{ki}  & \sim  \text{HalfNormal}(1)  \qquad \forall k = 1, \dots, 3 \qquad \forall i = 1, \dots, 2  \\
y_{n}         & \sim \prod_{k=1}^K \pi_k \text{MvNormal}(\boldsymbol \mu_k, \boldsymbol \Sigma_k)\,\; \quad \forall n = 1, \dots, N
\end{split}
\label{eqn:mixture-model}
\end{equation}

\begin{table*}[h!]
\centering
\ra{1}
\begin{tabular}{@{}rrrrr rrr c rrr@{}}\toprule
&&&&& \multicolumn{3}{c}{$N=100$} & \phantom{abc} & \multicolumn{3}{c}{$N=1000$} \\
\cmidrule{6-8} \cmidrule{10-12} 
&$N_\text{diff}$ & $N_\text{steps}$ & $N_\text{order}$ && $\bar{T}_\text{train}$ & $\bar{T}_\text{sample}$ & MSE && $\bar{T}_\text{train}$ & $\bar{T}_\text{sample}$ & MSE\\
\midrule
ADVI            &&&&& $6.84$ & $0.35$ & $8.49$ && $23.72$ & $0.37$ & $0.75$\\
DMVI&$50$ & $10$ & $1$ && ${51.37}$ & $4.33$ & ${0.86}$ && $297.47$ & $4.01$ & $0.77$\\
&$50$ & $10$ & $3$ && $57.43$ & $7.70$ & $0.89$ && ${288.42}$ & $7.48$ & ${0.77}$\\
&$50$ & $20$ & $1$ &&$77.23$ & $6.34$ & $0.87$ && $556.48$ & $6.35$ & $0.77$ \\
&$50$ & $20$ & $3$ && $91.47$ & $11.75$ & $0.90$ && $473.91$ & $11.16$ & $0.77$\\

&$100$ & $10$ & $1$ && $55.81$ & $5.78$ & $0.88$ && $361.54$ & $5.63$ & $0.77$\\
&$100$ & $10$ & $3$ && $69.59$ & $9.52$ & $0.88$ && $306.45$ & $8.47$ & $0.77$\\
&$100$ & $20$ & $1$ && $99.82$ & $9.86$ & $0.89$ && $639.02$ & $9.03$ & $0.77$\\
&$100$ & $20$ & $3$ && $96.03$ & $13.98$ & $0.89$ && $564.77$ & $13.37$ & $0.77$\\

NFVI             &&&&& $77.34$ & ${1.26}$ & $0.89$ && $440.61$ & ${1.17}$ & $0.85$\\
\bottomrule
\end{tabular}
\caption{Model of equation \ref{eqn:mixture-model}.}
\end{table*}

\subsection{Hierarchical model 1}

\begin{equation}
\begin{split}
\gamma_i        & \sim  \text{Normal}(0, 1) \;\; \quad \forall i = 1, 2 \\
\beta_{ij}      & \sim  \text{Normal}(\gamma_i, 1)  \qquad \forall j = 1, \dots, 5  \\
\sigma    & \sim  \text{HalfNormal}(1) \\
y_{ijn}     & \sim \text{Normal}(\beta_{ij}, \sigma^2) \,\; \quad \forall n = 1, \dots, N
\end{split}
\label{eqn:hierarchical-model1}
\end{equation}

\begin{table*}[h!]
\centering
\ra{1}
\begin{tabular}{@{}rrrrr rrr c rrr@{}}\toprule
&&&&& \multicolumn{3}{c}{$N=100$} & \phantom{abc} & \multicolumn{3}{c}{$N=1000$} \\
\cmidrule{6-8} \cmidrule{10-12} 
&$N_\text{diff}$ & $N_\text{steps}$ & $N_\text{order}$ && $\bar{T}_\text{train}$ & $\bar{T}_\text{sample}$ & MSE && $\bar{T}_\text{train}$ & $\bar{T}_\text{sample}$ & MSE\\
\midrule
ADVI &&&&& $6.77$ & $0.33$ & $4.80$ && $23.52$ & $0.33$ & $0.78$ \\

DMVI&$50$ & $10$ & $1$ && $52.72$ & $4.47$ & $0.19$ && $319.24$ & $4.42$ & $0.20$\\
&$50$ & $10$ & $3$ && $59.40$ & $7.76$ & $0.19$ && $281.85$ & $7.50$ & $0.20$\\
&$50$ & $20$ & $1$ && $85.75$ & $6.91$ & $0.19$ && $572.51$ & $6.89$ & $0.20$\\
&$50$ & $20$ & $3$ && $88.20$ & $11.72$ & $0.19$ && $487.05$ & $11.46$ & $0.20$\\

&$100$ & $10$ & $1$ && $58.08$ & $5.91$ & $0.19$ && $391.41$ & $5.97$ & $0.20$\\
&$100$ & $10$ & $3$ && $64.83$ & $9.45$ & $0.19$ && $370.02$ & $9.73$ & $0.20$\\
&$100$ & $20$ & $1$ && $105.10$ & $10.37$ & $0.19$ && $753.14$ & $10.39$ & $0.20$\\
&$100$ & $20$ & $3$ && $103.22$ & $15.18$ & $0.19$ && $604.08$ & $14.81$ & $0.20$\\

NFVI &&&&& $80.78$ & $1.20$ & $0.23$ && $459.85$ & $1.24$ & $0.22$\\
\bottomrule
\end{tabular}
\caption{Model of equation \ref{eqn:hierarchical-model1}.}
\end{table*}

\subsection{Hierarchical model 2}

\begin{equation}
\begin{split}
\mu_\gamma        & \sim  \text{Normal}(0, 1) \\
\gamma_i        & \sim  \text{Normal}(\mu_\gamma, 1) \;\; \quad \forall i = 1, 2 \\
\beta_{ij}      & \sim  \text{Normal}(\gamma_i, 1)  \qquad \forall j = 1, \dots, 5  \\
\sigma    & \sim  \text{HalfNormal}(1) \\
y_{ijn}     & \sim \text{Normal}(\beta_{ij}, \sigma^2) \,\; \quad \forall n = 1, \dots, N
\end{split}
\label{eqn:hierarchical-model2}
\end{equation}

\begin{table*}[h!]
\centering
\ra{1}
\begin{tabular}{@{}rrrrr rrr c rrr@{}}\toprule
&&&&& \multicolumn{3}{c}{$N=100$} & \phantom{abc} & \multicolumn{3}{c}{$N=1000$} \\
\cmidrule{6-8} \cmidrule{10-12} 
&$N_\text{diff}$ & $N_\text{steps}$ & $N_\text{order}$ && $\bar{T}_\text{train}$ & $\bar{T}_\text{sample}$ & MSE && $\bar{T}_\text{train}$ & $\bar{T}_\text{sample}$ & MSE\\
\midrule
ADVI &&&&& $7.29$ & $0.48$ & $5.91$ && $23.25$ & $0.50$ & $1.69$  \\

DMVI&$50$ & $10$ & $1$ && $55.20$ & $4.57$ & $0.56$ && $329.90$ & $4.37$ & $0.56$\\
&$50$ & $10$ & $3$ && $59.26$ & $7.91$ & $0.56$ && $302.78$ & $7.92$ & $0.56$ \\
&$50$ & $20$ & $1$ && $86.11$ & $7.04$ & $0.56$ && $603.93$ & $6.92$ & $0.56$\\
&$50$ & $20$ & $3$ && $94.60$ & $12.03$ & $0.56$ && $490.44$ & $11.49$ & $0.56$ \\

&$100$ & $10$ & $1$ && $61.05$ & $6.10$ & $0.56$ && $394.79$ & $6.15$ & $0.56$\\
&$100$ & $10$ & $3$ && $70.14$ & $10.03$ & $0.57$ && $354.46$ & $9.65$ & $0.56$\\
&$100$ & $20$ & $1$ && $109.24$ & $10.68$ & $0.56$ && $766.47$ & $10.78$ & $0.56$\\
&$100$ & $20$ & $3$ && $101.82$ & $14.96$ & $0.57$ && $625.96$ & $15.62$ & $0.56$\\

NFVI &&&&& $81.31$ & $1.20$ & $0.59$ && $552.86$ & $1.24$ & $0.58$\\
\bottomrule
\end{tabular}
\caption{Model of equation \ref{eqn:hierarchical-model2}.}
\end{table*}

\subsection{Hierarchical model 3}

\begin{equation}
\begin{split}
\sigma_\gamma   & \sim  \text{HalfNormal}(1) \\
\mu_\gamma      & \sim  \text{Normal}(0, 1) \\
\gamma_i        & \sim  \text{Normal}(\mu_\gamma, \sigma^2_\gamma) \;\; \quad \forall i = 1, 2 \\
\beta_{ij}      & \sim  \text{Normal}(\gamma_i, 1)  \qquad \forall j = 1, \dots, 5  \\
\sigma    & \sim  \text{HalfNormal}(1) \\
y_{ijn}     & \sim \text{Normal}(\beta_{ij}, \sigma^2) \,\; \quad \forall n = 1, \dots, N
\end{split}
\label{eqn:hierarchical-model3}
\end{equation}

\begin{table*}[h!]
\centering
\ra{1}
\begin{tabular}{@{}rrrrr rrr c rrr@{}}\toprule
&&&&& \multicolumn{3}{c}{$N=100$} & \phantom{abc} & \multicolumn{3}{c}{$N=1000$} \\
\cmidrule{6-8} \cmidrule{10-12} 
&$N_\text{diff}$ & $N_\text{steps}$ & $N_\text{order}$ && $\bar{T}_\text{train}$ & $\bar{T}_\text{sample}$ & MSE && $\bar{T}_\text{train}$ & $\bar{T}_\text{sample}$ & MSE\\
\midrule
ADVI &&&&& $7.56$ & $0.35$ & $5.94$ && $42.50$ & $0.39$ & $3.74$   \\

DMVI&$50$ & $10$ & $1$ && $59.20$ & $4.73$ & $0.77$ && $366.97$ & $4.60$ & $1.46$ \\
&$50$ & $10$ & $3$ &&  $60.04$ & $7.95$ & $0.69$ && $298.32$ & $8.01$ & $1.29$\\
&$50$ & $20$ & $1$ && $86.33$ & $7.13$ & $0.82$ && $594.04$ & $6.82$ & $1.48$\\
&$50$ & $20$ & $3$ && $95.32$ & $12.07$ & $0.70$ && $519.81$ & $12.14$ & $1.43$  \\

&$100$ & $10$ & $1$ && $63.37$ & $5.99$ & $0.71$ && $418.98$ & $6.13$ & $1.42$\\
&$100$ & $10$ & $3$ && $67.96$ & $9.71$ & $0.59$ && $388.37$ & $10.44$ & $1.31$\\
&$100$ & $20$ & $1$ && $103.01$ & $10.12$ & $0.67$ && $779.95$ & $10.52$ & $1.45$\\
&$100$ & $20$ & $3$ && $105.09$ & $15.20$ & $0.63$ && $664.93$ & $16.13$ & $1.37$\\

NFVI &&&&& $90.72$ & $1.32$ & $0.39$ && $484.22$ & $1.18$ & $5.13$\\
\bottomrule
\end{tabular}
\caption{Model of equation \ref{eqn:hierarchical-model3}.}
\end{table*}

\subsection{Hierarchical model 4}

\begin{equation}
\begin{split}
\sigma_\gamma   & \sim  \text{HalfNormal}(1) \\
\mu_\gamma      & \sim  \text{Normal}(0, 1) \\
\gamma_i        & \sim  \text{Normal}(\mu_\gamma, \sigma^2_\gamma) \;\; \quad \forall i = 1, 2 \\
\sigma_\beta    & \sim  \text{HalfNormal}(1) \\
\beta_{ij}      & \sim  \text{Normal}(\gamma_i, \sigma^2_\beta)  \qquad \forall j = 1, \dots, 5  \\
y_{ijn}     & \sim \text{Normal}(\beta_{ij}, 1) \,\; \quad \forall n = 1, \dots, N
\end{split}
\label{eqn:hierarchical-model4}
\end{equation}

\begin{table*}[h!]
\centering
\ra{1}
\begin{tabular}{@{}rrrrr rrr c rrr@{}}\toprule
&&&&& \multicolumn{3}{c}{$N=100$} & \phantom{abc} & \multicolumn{3}{c}{$N=1000$} \\
\cmidrule{6-8} \cmidrule{10-12} 
&$N_\text{diff}$ & $N_\text{steps}$ & $N_\text{order}$ && $\bar{T}_\text{train}$ & $\bar{T}_\text{sample}$ & MSE && $\bar{T}_\text{train}$ & $\bar{T}_\text{sample}$ & MSE\\
\midrule
ADVI &&&&& $7.74$ & $0.36$ & $12.47$ && $40.12$ & $0.36$ & $3.78$  \\

DMVI&$50$ & $10$ & $1$ && $50.73$ & $4.42$ & $0.63$ && $353.61$ & $4.85$ & $1.15$ \\
&$50$ & $10$ & $3$ &&  $59.55$ & $7.75$ & $0.63$    && $290.16$ & $7.81$ & $1.15$\\
&$50$ & $20$ & $1$ && $90.13$ & $7.11$ & $0.64$     && $595.12$ & $6.82$ & $1.12$\\
&$50$ & $20$ & $3$ && $89.73$ & $11.73$ & $0.63$    && $501.43$ & $11.70$ & $1.15$ \\

&$100$ & $10$ & $1$ && $58.67$ & $5.96$ & $0.63$ && $443.80$ & $6.55$ & $1.16$\\
&$100$ & $10$ & $3$ && $67.05$ & $9.67$ & $0.55$ && $337.37$ & $9.41$ & $1.12$\\
&$100$ & $20$ & $1$ && $102.74$ & $10.51$ & $0.63$ && $713.09$ & $10.27$ & $1.09$\\
&$100$ & $20$ & $3$ && $107.99$ & $15.53$ & $0.58$ && $600.31$ & $14.70$ & $1.16$\\

NFVI &&&&& $87.41$ & $1.25$ & $0.64$ && $555.45$ & $1.20$ & $4.23$\\
\bottomrule
\end{tabular}
\caption{Model of equation \ref{eqn:hierarchical-model4}.}
\end{table*}

\subsection{Hierarchical model 5}

\begin{equation}
\begin{split}
\sigma_\gamma   & \sim  \text{HalfNormal}(1) \\
\mu_\gamma      & \sim  \text{Normal}(0, 1) \\
\gamma_i        & \sim  \text{Normal}(\mu_\gamma, \sigma^2_\gamma) \;\; \quad \forall i = 1, \dots, 5 \\
\sigma_\beta    & \sim  \text{HalfNormal}(1) \\
\beta_{ij}      & \sim  \text{Normal}(\gamma_i, \sigma^2_\beta)  \qquad \forall j = 1, 2  \\
y_{ijn}     & \sim \text{Normal}(\beta_{ij}, 1) \,\; \quad \forall n = 1, \dots, N
\end{split}
\label{eqn:hierarchical-model5}
\end{equation}

\begin{table*}[h!]
\centering
\ra{1}
\begin{tabular}{@{}rrrrr rrr c rrr@{}}\toprule
&&&&& \multicolumn{3}{c}{$N=100$} & \phantom{abc} & \multicolumn{3}{c}{$N=1000$} \\
\cmidrule{6-8} \cmidrule{10-12} 
&$N_\text{diff}$ & $N_\text{steps}$ & $N_\text{order}$ && $\bar{T}_\text{train}$ & $\bar{T}_\text{sample}$ & MSE && $\bar{T}_\text{train}$ & $\bar{T}_\text{sample}$ & MSE\\
\midrule
ADVI &&&&& $7.38$ &	$0.37$ & $11.77$                     && $38.87$ & $0.40$ & $2.54$\\

DMVI&$50$ & $10$ & $1$ && $57.18$ &	$4.69$  & $0.49$                 && $276.22$ &	$3.93$ & $1.09$\\
&$50$ & $10$ & $3$ && ${54.52}$  & $7.12$ & $0.46$        && ${266.37}$ &	$7.23$ & $1.07$\\
&$50$ & $20$ & $1$ && $92.67$ &	$7.70$  & $0.51$                 && $546.03$ &	$6.85$ & $1.10$\\
&$50$ & $20$ & $3$ && $83.29$ &	$10.83$ & $0.52$                 && $491.99$ &	$11.22$ & $1.05$\\

&$100$ & $10$ & $1$ && $66.07$  & $6.29$   & $0.41$             && $437.52$ & $6.27$  & $2.22$\\
&$100$ & $10$ & $3$ && $64.50$  & $9.22$   & ${0.39}$    && $353.74$ & $9.18$ & ${1.03}$\\
&$100$ & $20$ & $1$ && $110.28$ & $10.55$  & $0.40$             && $757.28$ & $9.78$  & $3.47$\\
&$100$ & $20$ & $3$ && $107.82$ & $15.53$  & $0.44$             && $605.38$ & $14.38$ & $1.03$\\

NFVI &&&&& $96.01$ 	& ${1.17}$ &$0.62$                     && $689.78$ & ${1.32}$ & $3.99$\\
\bottomrule
\end{tabular}
\caption{Model of equation \ref{eqn:hierarchical-model5}.}
\end{table*}

\end{appendices}

\end{document}